\documentclass{article}
\usepackage{spconf,amsmath,graphicx}
\usepackage{multirow}
\usepackage{cite}
\usepackage{booktabs}
\usepackage{amssymb}
\usepackage{pdfpages}
\usepackage{CJKutf8}  

\usepackage{xcolor}
\usepackage{hyperref}

\title{Mandarin-English Code-switching Speech Recognition\\with Self-supervised Speech Representation Models}
\name{Liang-Hsuan Tseng$^{\ast1}$, Yu-Kuan Fu$^{\ast2}$, Heng-Jui Chang$^1$, Hung-yi Lee$^1$ \thanks{$^{\ast}$ Equal contribution.}}
\address{
$^1$Department of Electrical Engineering, National Taiwan University\\
$^2$Department of Physics, National Taiwan University
}
\begin{document}
\ninept
\maketitle
\begin{abstract}
Code-switching (CS) is common in daily conversations where more than one language is used within a sentence.
The difficulties of CS speech recognition lie in alternating languages and the lack of transcribed data.
Therefore, this paper uses the recently successful self-supervised learning (SSL) methods to leverage many unlabeled speech data without CS.
We show that hidden representations of SSL models offer frame-level language identity even if the models are trained with English speech only.
Jointly training CTC and language identification modules with self-supervised speech representations improves CS speech recognition performance.
Furthermore, using multilingual speech data for pre-training obtains the best CS speech recognition.
\end{abstract}

\section{Introduction}
\label{sec:intro}

Code-switching (CS) refers to speaking two or more languages in an utterance.
CS speech is frequently used in daily conversations in areas using multiple languages like Southeast Asia; therefore, recognizing this kind of speech is crucial.
However, transcribed CS speech data are uneasy to collect, and the unique property of switching different languages within a sentence increases machine recognition difficulty.

Studies show deep learning-based end-to-end automatic speech recognition (ASR) methods like connectionist temporal classification (CTC) \cite{graves2014ctcasr} successful in various speech recognition tasks.
Unlike conventional ASR technologies, end-to-end ASR methods jointly optimize all model parameters, offering better performance and higher flexibility.
Building on top of end-to-end ASR technologies, researchers proposed many approaches to tackle the CS speech recognition problem, including predicting language identity \cite{li2019towards,shan2019investigating,zeng2019end,zhang2021rnn}, leveraging monolingual data \cite{shan2019investigating,lu2020bi,zhou2020multi,dalmia2021transformer,chuang2021non}, and data augmentation techniques \cite{chang2019code,long2020acoustic,sharma2020improving,du2021data}.

A recently successful method to tackle the data scarcity problem in speech processing tasks is using self-supervised learning (SSL) pre-trained models.
SSL is a machine learning method by learning to predict pseudo labels derived from unlabeled data.
Using SSL models to learn good speech representations succeeded in various downstream tasks like ASR, speaker identification, and emotion recognition \cite{yang2021superb}.
These speech SSL methods can be roughly categorized into generative \cite{liu2020mockingjay,chung2019apc,chung2020vq-apc,liu2021tera,liu2021npc} and discriminative \cite{oord2018cpc,riviere2020m-cpc,schneider2019wav2vec,baevski2020vq-wav2vec,baevski2020wav2vec2,hsu2021hubert} training objectives.

To our knowledge, this is the first paper investigating the effectiveness of self-supervised speech representation models for CS speech recognition and the capabilities of these models offering frame-wise language identity information.
First, we showed that the SSL models' hidden representations contained language identity.
Then, the SSL models improved Mandarin-English CS speech recognition, and even they were pre-trained with English data only.
Moreover, incorporating the language identification task with CTC ASR further boosted CS speech recognition.

\section{Methods}
\label{sec:method}

\subsection{wav2vec 2.0}
\label{ssec:wav2vec2}

This paper investigates the effectiveness of self-supervised models on the Mandarin-English CS speech recognition task, including comparing monolingual and multilingual SSL models.
To achieve this goal, we found wav2vec 2.0 \cite{baevski2020wav2vec2} was the most suitable method to use in our experiments, which is an SSL framework trained with contrastive learning, consisting of a 7-layer CNN feature extractor and a transformer encoder \cite{vaswani2017attention}.
First, most publicly available models are trained with English speech, while wav2vec 2.0 has many variations trained with different datasets, including a multilingual model.
Next, studies show successful results on numerous end-to-end ASR tasks using hidden representations of wav2vec 2.0 \cite{baevski2020wav2vec2,pasad2021layer,chang2021exploration,yang2021superb}.

\subsection{Language Identification by SSL Models}
\label{ssec:LID}

We wish to investigate whether pre-trained speech representation models offer language-related information because identifying the languages in utterances is critical in CS speech recognition.
A simple task to examine this fact is language identification (LID).
LID is a task to determine the language of a given speech signal segment.
Many works investigate utterance-level LID \cite{jin2017end,cai2019utterance}, where each utterance contains only one language.
In CS speech, at least two languages occur in each utterance.
Hence, LID in this paper refers to frame-wise LID, which is a sequence labeling problem, aiming to classify each time frame in an utterance into either silence or a language.

Some previous studies in CS speech recognition obtain good results on LID using acoustic features or context features extracted from ASR models \cite{li2019towards,shan2019investigating,zeng2019end,zhang2021rnn}.
In contrast, this paper is the first attempt to perform LID using SSL pre-trained models.
Since studies show wav2vec 2.0 hidden representations contain semantic or word meaning information \cite{baevski2020wav2vec2}, we wish to verify whether they carry language identity.

\begin{figure}[t]
    \centering
    \includegraphics[width=0.85\linewidth]{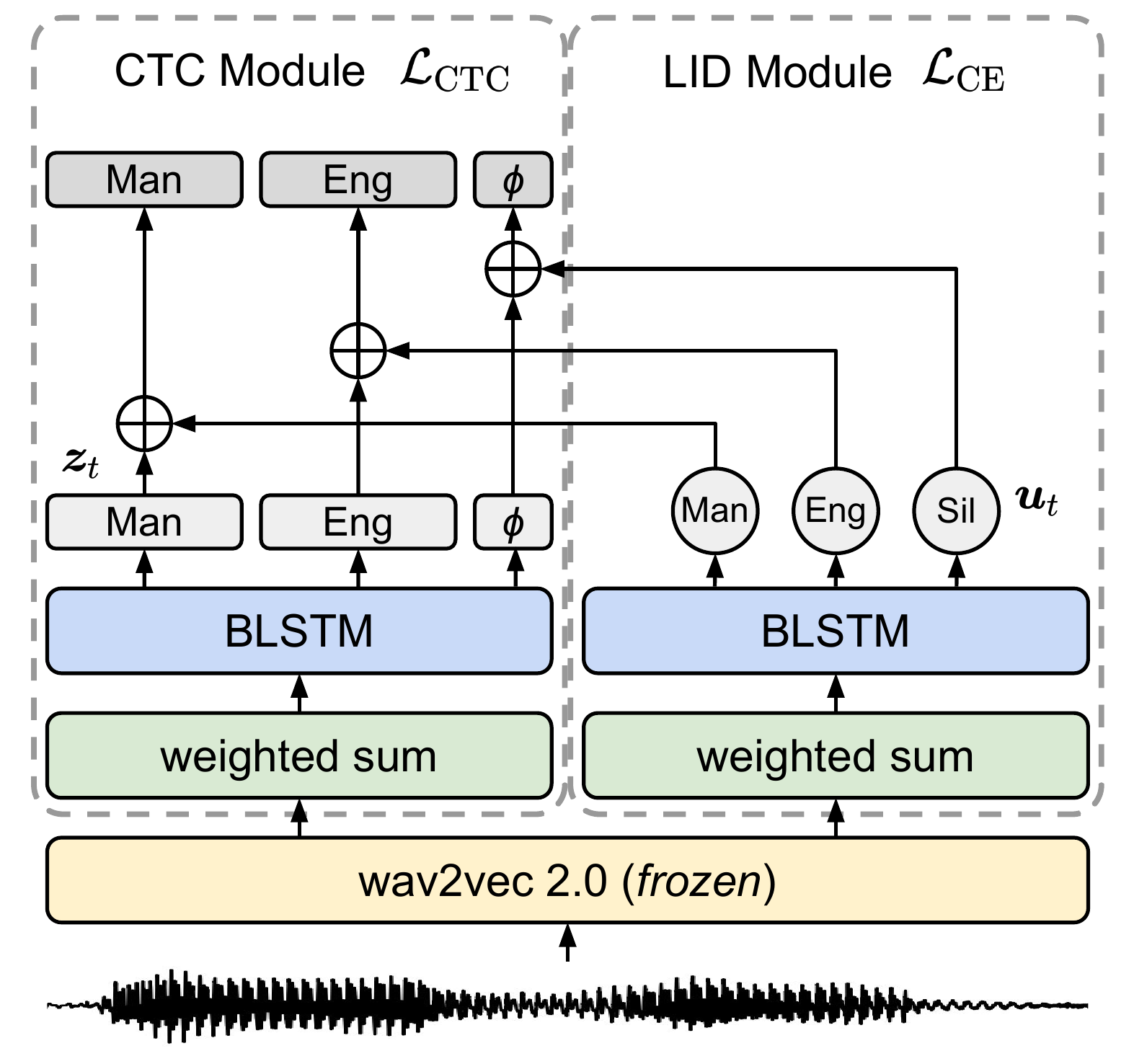}
    \caption{
        The joint CTC-LID ASR framework.
        The left and right parts are respectively the CTC and LID modules.
        Man, Eng, and $\phi$ respectively denote Mandarin, English, and CTC's blank token.
    }
    \label{fig:model}
\end{figure}

\subsection{Joint CTC-LID Training}
\label{ssec:plid}

In this section, we propose a joint CTC-LID framework similar to Li et al. \cite{li2019towards} for end-to-end CS speech recognition.
However, the main difference is that the input features are hidden representations of a pre-trained SSL model, as shown in Fig. \ref{fig:model}.
This framework transfers the burden of identifying the CS phenomenon from the ASR model to an additional LID module.

The joint CTC-LID framework consists of three parts: feature extraction, CTC, and LID modules.
First, the input waveform is passed through wav2vec 2.0 to extract good speech representations.
Next, two trainable weighted sum mechanisms summarize the hidden layers' representations of wav2vec 2.0.
Then, the two summarized sequences are respectively fed into the CTC and LID modules.
The LID module aims to classify each frame into either silence, Mandarin, or English, where the labels are obtained by forced alignment.
The CTC module is slightly different from typical CTC models \cite{graves2014ctcasr}; instead, the logits of all possible output tokens are scaled by the LID module's output.
This mechanism shares the information between the CTC and LID modules, and we wish it mitigates the LID task from CTC.

As shown in Fig. \ref{fig:model}, the Mandarin, English, and blank tokens' logits from the CTC module are respectively added with the Mandarin, English, and silence logits output by the LID module.
The new probability of each possible token $y$ in vocabulary $\mathcal{V}$ at frame $t$ given input features $X$ is
\begin{equation}
    P(y | X, t) = \frac{ \exp \left( \boldsymbol{z}_{t, y} + \boldsymbol{u}_{t, l(y)} \right) }{ \sum_{y'\in\mathcal{V}} \exp \left( \boldsymbol{z}_{t, y'} + \boldsymbol{u}_{t, l(y')} \right) },
\end{equation}
where $\boldsymbol{z}$ and $\boldsymbol{u}$ are respectively the output logits of the CTC and LID modules.
$l: \mathcal{V} \rightarrow $ \{Mandarin, English, $\phi$\} is a function returning the token type.
After scaling with logits $\boldsymbol{u}$, the loss function is
\begin{equation}
    \mathcal{L} = (1 - \lambda)\mathcal{L}_{\mathrm{CTC}} + \lambda \mathcal{L}_{\mathrm{CE}},
    \label{eq:loss}
\end{equation}
where $\lambda$ controls the contribution of the cross entropy loss.

\section{Experiments}
\label{sec:exp}

\begin{table}[t]
    \caption{The duration and the proportion of Mandarin, English, and code-switching utterances in the training/validation/testing sets of SEAME \cite{lyu2010seame}.}
    \centering
    \vspace{4pt}
    \begin{tabular}{lcccc}
        \toprule 
        & train & val & dev-man & dev-sge \\
        \midrule
        Duration (hours) & 93.0 & 4.7 & 7.5 & 4.0 \\
        Mandarin & 23.4\% & 23.7\% & 23.5\% & 9.8\% \\
        English & 21.4\% & 21.1\% & 11.2\% & 48.7\% \\
        Code-switching & 55.1\% & 55.2\% & 65.3\% & 41.5\% \\
        \bottomrule
    \end{tabular}
    \label{table:corpus}
\end{table}
\begin{figure}[t]
    \centering
    \includegraphics[width=\linewidth]{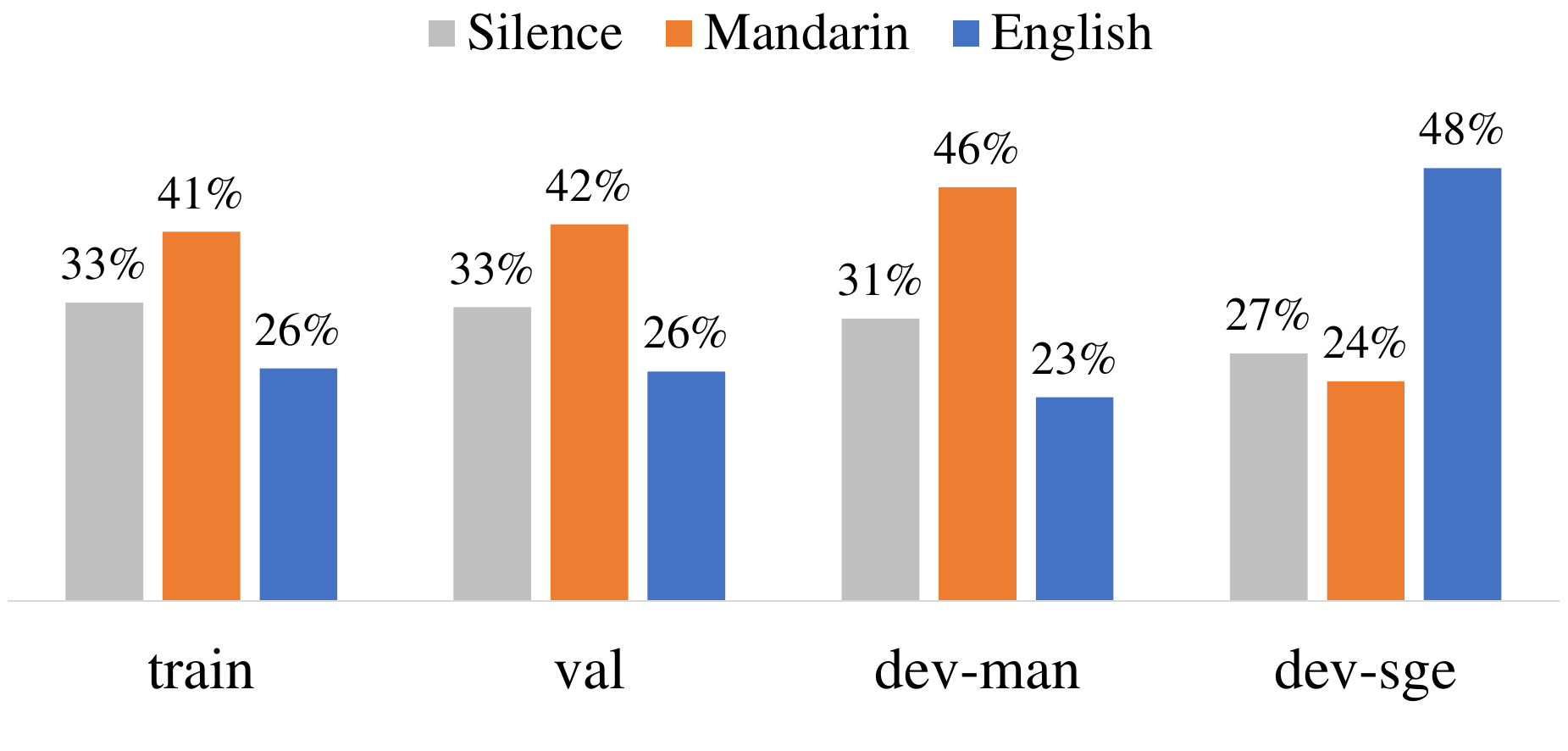}
    \vspace*{-20pt}
    \caption{The distribution of LID labels in each set.}
    \label{figure:lid_labels}
\end{figure}

\begin{table}[t]
    \caption{The wav2vec 2.0 models used in the experiments.}
    \centering
    \vspace{2pt}
    \begin{tabular}{lcccc}
        \toprule
        Model & Layers & Dim. & Data (hours) & Languages \\
        \midrule
        Base & 12 & 768 & 960 & 1 (English) \\
        Large & 24 & 1024 & 60k & 1 (English) \\
        XLSR & 24 & 1024 & 56k & 53 \\
        \bottomrule
    \end{tabular}
    \label{table:wav2vec2}
\end{table}

\subsection{Setup}
\label{exp:setup}

\noindent\textbf{Data.}
In the following experiments, we used the SEAME speech corpus \cite{lyu2010seame}, a Mandarin-English CS conversational dataset.
This corpus had two evaluation sets: dev-man and dev-sge, respectively biased towards Mandarin and English.
The composition of data is described in Table \ref{table:corpus}.
Following previous works in CS speech recognition \cite{zeng2019end,shan2019investigating,chuang2021non}, the vocabulary was composed of 2718 Mandarin characters and 2278 English subword units \cite{sennrich2015neural}.
The evaluation metric was token error rate (TER), where the tokens referred to Mandarin characters and English words.
The frame-wise LID task used Montreal Force Aligner \cite{mcauliffe2017montreal} to find word boundaries in each utterance.
Then, each frame was labeled either silence, Mandarin, or English.
The proportion of these labels in SEAME is shown in Fig. \ref{figure:lid_labels}.

\noindent\textbf{wav2vec 2.0.}
The baseline method without SSL pre-training used 80-dimensional filterbank (fbank) features with delta and delta-delta, resulting in a total dimension of 240.
The SSL pre-trained method here was wav2vec 2.0 \cite{baevski2020wav2vec2}, the models are listed in Table \ref{table:wav2vec2}.
The wav2vec 2.0 models were frozen throughout the experiments in this paper.
The hidden feature vectors of each layer's output of wav2vec 2.0 were normalized to zero-mean and unit variance, then summarized with a trainable weighted sum mechanism.

\noindent\textbf{LID \& CTC Models.}
The LID experiments in Sec. \ref{exp:lid} used two modules: a single fully-connected (FC) layer and a 1-layer BLSTM with a hidden dimension of 1024 per direction.
The joint CTC-LID model's CTC module had a 2-layer BLSTM with a hidden dimension of 1024 per direction, and the LID module was the BLSTM module as described earlier.
The constant $\lambda=$ 0.1 in Eq. (\ref{eq:loss}).
The baseline CTC model had a hidden dimension of 1216 to match the size of the joint CTC-LID model for fair comparisons.
All ASR experiments used SpecAugment \cite{park2019specaug}.

\begin{table}[t]
    \caption{Accuracy (\%) of LID using different methods.}
    \centering
    \vspace{2pt}
    \begin{tabular}{lcc}
        \toprule
        Method & dev-man & dev-sge \\
        \midrule
        (a) fbank + FC    & 59.4 & 37.7 \\
        (b) fbank + BLSTM & 84.7 & 82.3  \\
        \midrule
        (c) Base + FC     & 75.0 & 68.5 \\
        (d) Base + BLSTM  & 91.9 & 89.5 \\
        \midrule
        (e) Large + FC    & 77.9 & 71.8 \\
        (f) Large + BLSTM & 92.3 & 89.9 \\
        \midrule
        (g) XLSR + FC     & 76.4 & 69.7 \\
        (h) XLSR + BLSTM  & \textbf{92.7} & \textbf{90.0} \\
        \bottomrule
    \end{tabular}
    \label{table:lid}
\end{table}

\subsection{Language Identification}
\label{exp:lid}
We inspected the language identity information extracted from fbank and wav2vec 2.0 representations by performing frame-wise LID on SEAME.
The results are shown in Table \ref{table:lid}.
Using a simple FC layer as the prediction head, fbank features performed significantly worse than other wav2vec 2.0 models (rows (a) vs. (c)(e)(g)), showing that pre-trained speech representation models offered more language identity information.
Using a BLSTM model as the prediction heads captured better context information and improved LID accuracy in both fbank and wav2vec 2.0 (rows (b)(d)(f)(h)).
We showed that although wav2vec 2.0 Base and Large were trained solely on English speech, they could distinguish different languages within each sentence.
Moreover, we found wav2vec 2.0 Large and XLSR offered comparable performance in LID, perhaps because the characteristics of Mandarin and English speech were very different, making the models easy to determine even if the pre-trained model was trained with English.
Overall, wav2vec 2.0 had the potential to achieve better CS speech recognition than fbank.

\begin{table}[t]
    \caption{TERs (\%) of CS speech recognition, where FT in section (V) denotes fine-tuning.}
    \centering
    \vspace{2pt}
    \small
    \begin{tabular}{lcccccccc}
        \toprule
        & \multicolumn{3}{c}{dev-man} &
        \multicolumn{3}{c}{dev-sge} \\
        \cmidrule(r{4pt}){2-4} \cmidrule(l){5-7}
        Method & All & Man & Eng & All & Man & Eng \\
        \midrule
        \multicolumn{7}{l}{\textit{(I) Baseline}} \\
        \midrule
        (a) fbank       & 29.4 & 25.6 & 48.8 & 40.8 & 35.7 & 47.9 \\
        (b) fbank + LID & 29.0 & 25.7 & 47.2 & 40.6 & 38.7 & 47.5 \\
        \midrule
        \multicolumn{7}{l}{\textit{(II) wav2vec 2.0 + CTC}} \\
        \midrule
        (c) Base  & 20.8 & 18.1 & 33.5 & 30.8 & 26.9 & 36.3 \\
        (d) Large & 19.8 & 17.1 & 32.1 & 29.7 & 25.4 & 35.3 \\
        (e) XLSR  & 19.4 & 16.5 & 32.4 & 28.8 & \textbf{23.7} & 34.5 \\
        \midrule
        \multicolumn{7}{l}{\textit{(III) wav2vec 2.0 + CTC-LID (trained separately) \cite{li2019towards}}} \\
        \midrule
        (f) Base  & 21.0 & 18.4 & 33.9 & 31.4 & 27.1 & 37.3 \\
        (g) Large & 20.9 & 18.0 & 34.0 & 31.3 & 27.2 & 36.8 \\
        (h) XLSR  & 20.2 & 17.3 & 33.4 & 29.9 & 25.0 & 35.7 \\
        \midrule
        \multicolumn{7}{l}{\textit{(IV) wav2vec 2.0 + CTC-LID (jointly trained from scratch)}} \\
        \midrule
        (i) Base  & 20.8 & 18.2 & 33.6 & 30.9 & 27.6 & 36.3 \\
        (j) Large & 20.1 & 17.6 & 32.7 & 30.3 & 27.3 & 35.8 \\
        (k) XLSR  & 19.2 & 16.5 & 31.9 & 29.1 & 25.0 & 34.5 \\
        \midrule
        \multicolumn{7}{l}{\textit{(V) wav2vec 2.0 + CTC-LID (trained separately + jointly FT)}} \\
        \midrule
        (l) Base  & 20.3 & 17.7 & 33.1 & 30.6 & 26.6 & 36.2 \\
        (m) Large & 19.9 & 17.3 & 32.4 & 29.8 & 25.5 & 35.3 \\
        (n) XLSR  & \textbf{18.8} & \textbf{16.1} & \textbf{30.8} & \textbf{28.5} & 23.8 & \textbf{34.1} \\
        \bottomrule
    \end{tabular}
    \label{table:all}
\end{table}

\subsection{Code-switching Speech Recognition}
\label{exp:cssr}
We examined the benefits of pre-trained speech representation models for CS speech recognition.
The results are shown in Table \ref{table:all}.
The baseline experiments used fbank features (section (I)).
The jointly trained CTC-LID reduced the TERs on the two evaluation sets by roughly 1\% (rows (b) vs. (a)), showing LID offered minor improvements because fbank had little language identity information as shown in the previous section.

Next, we discussed the results using wav2vec 2.0 as the CTC models' inputs (section (II)).
Comparing with the fbank baseline, wav2vec 2.0 offered a relative 24\% to 34\% TER reduction on both testing sets (rows (c)(d)(e) vs. (a)).
Furthermore, the results showed that using a larger SSL model performed better (rows (d) vs. (c)), even it was pre-trained with English speech data only.
The multilingual XLSR model offered the best overall performance (rows (e) vs. (c)(d)), indicating that including Mandarin speech into pre-training of wav2vec 2.0 alleviated the language mismatch in Mandarin-English CS speech.
We found a slight degradation in English recognition of the dev-man set because pre-training with other languages increased the capability of wav2vec 2.0 processing Mandarin speech but sacrificed English.
To this end, the findings showed pre-training with unlabeled speech data was beneficial for CS speech recognition, even the languages were mismatched.

Here, we investigated the effectiveness of the LID module for CTC ASR models.
First, we tried the methods proposed by Li et al. \cite{li2019towards}, i.e., the CTC and LID models were trained separately, and the CTC output probabilities were scaled by the LID probabilities directly (section (III)).
The results showed slight degradation on both evaluation sets, indicating this method was harmful to wav2vec 2.0.
This phenomenon was probably caused by the fact that CTC had no forced alignments between the output sequence and the input acoustic features \cite{senior2015acoustic}.
Hence, the CTC and LID models' predictions were unlikely to be aligned together.

Moreover, we tried training CTC and LID modules jointly from scratch (section (IV)), but this method showed similar performance compared with the case when only the CTC model was used (sections (IV) vs. (II)).
We explained that the jointly training framework made the CTC model learning process noisy and unstable.
Since LID offered erroneous labels at the beginning of training, the bad initialization damaged further CTC training.

Finally, in section (V), we proposed fine-tuning pre-trained CTC and LID modules jointly.
The results showed further improvements compared to other methods using SSL models (sections (V) vs. (II)(III)(IV)), indicating a good initialization of the two modules was critical in the joint CTC-LID framework.
Overall, self-supervised speech representation models benefited CS speech recognition, especially when the model was pre-trained with multilingual speech.

\subsection{Benefits of LID Brought to CTC}
\begin{figure}[t]
    \centering
    \includegraphics[width=1.0\linewidth]{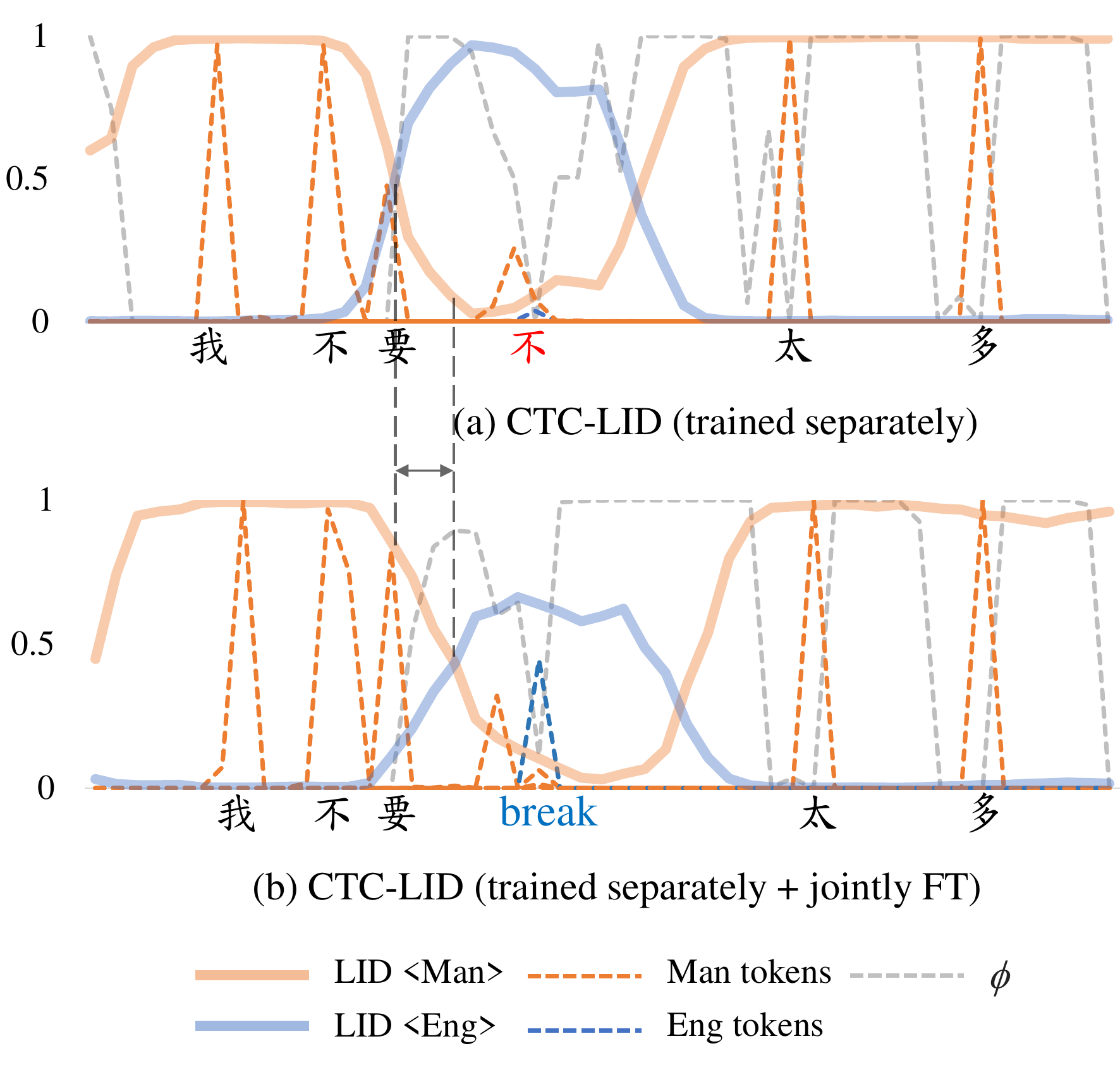}
    \caption{
        The posteriors of CTC and LID model.
        (a) The ground truth of the red word was "break", but was misspelled as the Mandarin character "\begin{CJK*}{UTF8}{bkai}不\end{CJK*}(bu)."
        (b) With the proposed jointly fine-tuning method ((V) of Table \ref{table:all}), the logits of Mandarin language identity helped decoding the character "\begin{CJK*}{UTF8}{bkai}不\end{CJK*}" correctly.
    }
    \label{fig:ctc_lid_plot}
\end{figure}
\begin{figure}[t]
    \centering
    \includegraphics[width=1.4\linewidth]{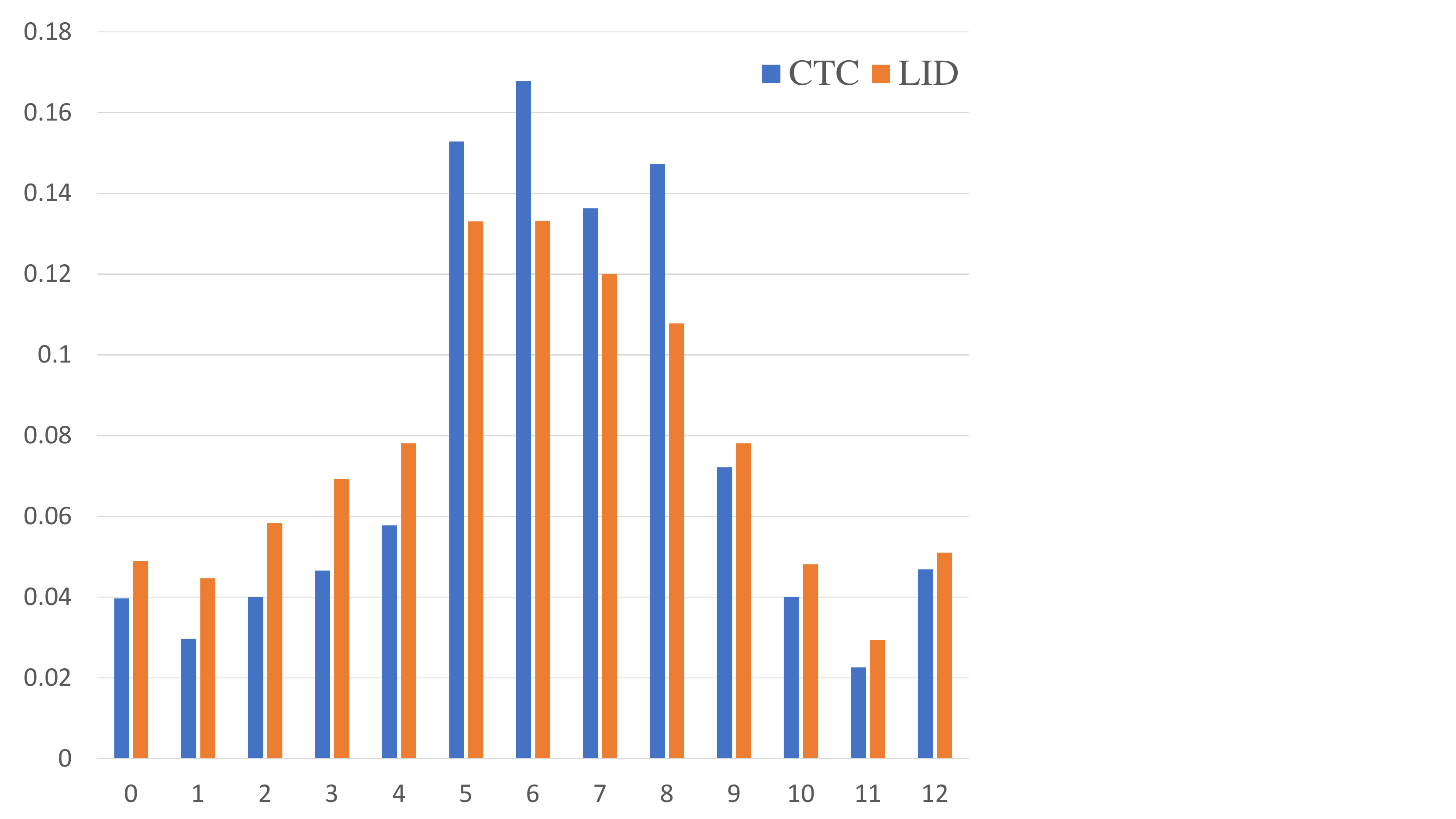}
    \caption{
        The weights for summarizing all layers' hidden representations of wav2vec 2.0 Base for CTC and LID.
        The weights are normalized by the averaged $\ell2$ norm of each layer's representations.
    }
    \label{fig:weights}
\end{figure}
In this section, we took a deeper look at how LID helped CTC decoding.
We selected one utterance in the dev-man set as an example and recognized it with two CTC-LID training strategies: trained separately \cite{li2019towards} and fine-tuned jointly (rows (f)(l) in Table \ref{table:all}).
The probabilities of the CTC and LID outputs are shown in Fig. \ref{fig:ctc_lid_plot}.
The two methods in Fig. \ref{fig:ctc_lid_plot} were quite different.
The former method directly multiplied the probabilities of CTC and LID, while the latter summed the logits together to fuse the two modules' outputs.
Because showing the logits was difficult to compare the two methods, we depicted the probabilities instead.

First, when CTC and LID were trained separately and used the decoding strategy proposed in \cite{li2019towards} (Fig. \ref{fig:ctc_lid_plot}(a)), the Mandarin probabilities from the LID module was low when CTC was emitting the character "\begin{CJK*}{UTF8}{bkai}要\end{CJK*}(yao)," suppressing the probability of this character.
Because the two modules were trained separately, the CTC and LID output probabilities were unaligned when code-switching occurred.

Then, when fine-tuning the two pre-trained modules jointly (Fig. \ref{fig:ctc_lid_plot}(b)), the LID module had similar output probabilities ((b) vs. (a)).
The main difference was at the location when CTC emitted "\begin{CJK*}{UTF8}{bkai}要\end{CJK*}," the LID probability of Mandarin was higher than the previous method, showing that the CTC and LID modules were aligned together.
Moreover, the jointly fine-tuning framework benefited CTC decoding since the English word "break" was correctly decoded.
In contrast, the previous method decoded it as the Mandarin character "\begin{CJK*}{UTF8}{bkai}不\end{CJK*}(bu)."
Although this example was only one of the decoding results, we found this phenomenon frequently occurred, showing that the LID module was beneficial to CTC.

\subsection{Layer Importance for CTC and LID}
To further investigate the importance of each hidden layer of wav2vec 2.0 to ASR and LID, we depicted the normalized weights of the weighted sum mechanism for CTC and LID.
The weights of the separately trained modules are shown in Fig. \ref{fig:weights} (same models from row (f) of Table \ref{table:all}), where higher values indicate greater importance of the layer.

The weights distribution of the two tasks were similar, showing that CTC and LID tasks relied on similar hidden features of wav2vec 2.0.
This phenomenon implied that the model relied on contextual information to identify languages, which behaved similarly to CS speech recognition.
Hence, as expected, the CS speech recognition task involved identifying languages simultaneously.
The joint CTC-LID training framework exploited this property to transfer the burden of identifying language from CTC to another module, resulting in better performance on CS speech recognition.

\section{Conclusion}
\label{sec:conclusion}

This paper proposes using self-supervised pre-trained speech representation models to tackle the code-switching speech recognition problem.
We showed that these models offered language identity and improved CS speech recognition accuracy.
Using an additional LID module further boosted performance.
Moreover, pre-training SSL models with multilingual data obtained better CS speech recognition.
In future works, we wish to investigate other benefits brought from other SSL models.

\bibliographystyle{IEEEtran}
\bibliography{refs,cs,ssl}

\end{document}